\documentclass[10pt,twocolumn,letterpaper]{article}

\usepackage[pagenumbers]{wacv} 

\usepackage{graphicx}
\usepackage{booktabs}
\usepackage{amsfonts}
\usepackage{makecell}
\usepackage{amssymb}
\usepackage{multirow}
\usepackage{float}
\usepackage{amsmath} 
\usepackage[pagebackref,breaklinks,colorlinks]{hyperref}

\usepackage[capitalize]{cleveref}
\crefname{section}{Sec.}{Secs.}
\Crefname{section}{Section}{Sections}
\Crefname{table}{Table}{Tables}
\crefname{table}{Tab.}{Tabs.}

\def\confName{WACV}
\def\confYear{2025}

\begin{document}

\title{Global-guided Focal Neural Radiance Field for Large-scale Scene Rendering}
\author{
Mingqi Shao$^{\dag1,2}$ \and
Feng Xiong$^{\dag2}$  \and
Hang Zhang$^2$ \and
Shuang Yang$^2$ \and
Mu Xu$^2$ \and
Wei Bian$^2$ \and
Xueqian Wang*$^1$ \\ 
$^1$Tsinghua Shenzhen International Graduate School \\
$^2$AMAP\\
{\tt\small smq21@mails.tsinghua.edu.cn~wang.xq@sz.tsinghua.edu.cn} \\
{\tt\small \{xf250971,suishou.zh,pingan.ys,xumu.xm,bianwei.ba\}@alibaba-inc.com~}
}

\maketitle
\footnotetext{$\dag$~These authors contributed equally to this work}
\footnotetext{*~Corresponding Author}
\begin{abstract}
Neural radiance fields~(NeRF) have recently been applied to render large-scale scenes. However, their limited model capacity typically results in blurred rendering results. Existing large-scale NeRFs primarily address this limitation by partitioning the scene into blocks, which are subsequently handled by separate sub-NeRFs. These sub-NeRFs, trained from scratch and processed independently, lead to inconsistencies in geometry and appearance across the scene. Consequently, the rendering quality fails to exhibit significant improvement despite the expansion of model capacity. In this work, we present global-guided focal neural radiance field (GF-NeRF) that achieves high-fidelity rendering of large-scale scenes. Our proposed GF-NeRF utilizes a two-stage (Global and Focal) architecture and a global-guided training strategy. The global stage obtains a continuous representation of the entire scene while the focal stage decomposes the scene into multiple blocks and further processes them with distinct sub-encoders. Leveraging this two-stage architecture, sub-encoders only need fine-tuning based on the global encoder, thus reducing training complexity in the focal stage while maintaining scene-wide consistency. Spatial information and error information from the global stage also benefit the sub-encoders to focus on crucial areas and effectively capture more details of large-scale scenes. Notably, our approach does not rely on any prior knowledge about the target scene, attributing GF-NeRF adaptable to various large-scale scene types, including street-view and aerial-view scenes. We demonstrate that our method achieves high-fidelity, natural rendering results on various types of large-scale datasets. Our project page: \href{https://shaomq2187.github.io/GF-NeRF/}{https://shaomq2187.github.io/GF-NeRF/}

\end{abstract}

\section{Introduction}

The photo-realistic rendering of large-scale scenes has garnered significant research interest, particularly with the advent of neural radiance fields~(NeRF), which has demonstrated remarkable performance and simplicity~\cite{mildenhall2021nerf}. Such capability enables several important practical applications, including autonomous driving simulation~\cite{yang2023unisim,guo2023streetsurf,li2019aads,yang2020surfelgan,ost2021neural}, AR/VR applications~\cite{song2023nerfplayer,deng2022fov}, and 3D map\cite{tancik2022block,rematas2022urban}. However, applying the original NeRF to large-scale scenes typically leads to significant artifacts and blurred renderings due to limited model capacity. Recent Mip-NeRF 360~\cite{barron2022mip} and F2-NeRF~\cite{wang2023f2} have enhanced NeRF's representational capabilities through space contraction. Mip-NeRF 360 warps unbounded scenes into bounded ranges, focusing computational resources on the observer's vicinity. F2-NeRF employs a perspective warping method to refine the hash grid encoder's efficiency, capturing more scene details. Despite these improvements, the models' capacity remains finite, leading to performance bottlenecks with larger scenes.

\begin{figure}[t]
\centering 
\includegraphics[width=\linewidth]{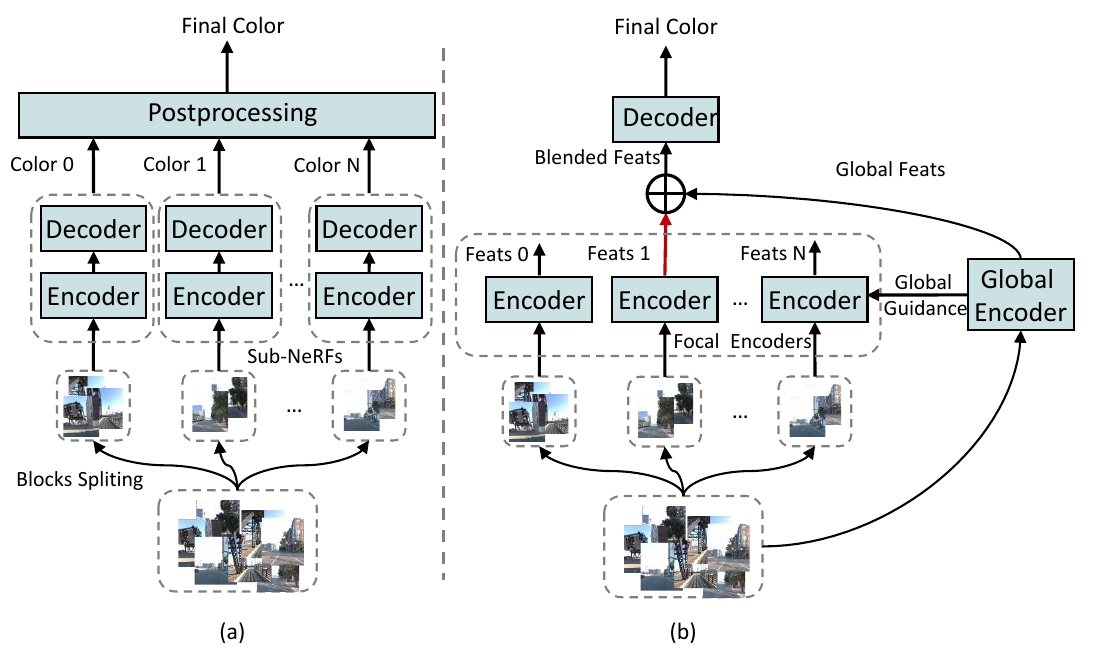} 
\caption{Different approaches of expanding NeRF capacity for large-scale scenes. (a) Expanding model capacity with independent sub-NeRFs and blending their colors through post-processing. (b) Introducing global information to guide the training of multiple blocks, expanding capacity while maintaining global consistency.} 
\label{fig:intro} 
\end{figure}

Recent research adopts the divide-and-conquer paradigm to NeRF~\cite{tancik2022block, turki2022mega}, partitioning scenes into blocks and training separate sub-NeRFs, which are then blended post-hoc, as shown in Fig. \ref{fig:intro} (a). This approach effectively extends NeRF's capacity but faces challenges with hand-crafted blending rules, leading to possible inconsistency among different sub-NeRFs. Switch-NeRF~\cite{zhenxing2022switch} mitigates this issue with a learnable gated mixture-of-experts model, at the cost of increased training complexity and potential blurry rendering results. Another coarse-to-fine idea, exemplified by Bungee-NeRF~\cite{xiangli2022bungeenerf}, enables progressive scene refinement. It adopts a hierarchical network structure from coarse to fine, with a base network for coarse scene representation, followed by residual blocks that progressively refine the details. While, this approach is tailored for extreme multi-scale scene rendering, particularly for the remote scene, and the coarse and fine stages operate independently, without leveraging prior insights from the coarse stage to inform the fine stage.

In this paper, we integrate the coarse-to-fine idea into the divide-and-conquer paradigm and propose the Global-guided Focal NeRF~(GF-NeRF) for large-scale scenes rendering. The key insight of our method is that the coarse stage within the coarse-to-fine encapsulates crucial global information about large-scale scenes which can be utilized to guide the training of the subsequent blocks. We partition the training of the large-scale scenes NeRF into two stages. In the first stage, referred to as the global stage, we employ a hash encoder\cite{muller2022instant} to initialize the entire scene. The primary objective during this stage is to acquire a coarse-grained spatial representation. In the second stage, referred to as the focal stage, we partition the scene into blocks and employ new hash encoders to learn residual features for fine-grained spatial representations as shown in Fig. \ref{fig:intro}~(b), which expands the overall NeRF's capacity while maintaining the consistency between blocks. Furthermore, the spatial information and error information from the global stage guide the training in the focal stage, forcing the second stage to focus on the areas that the global stage did not handle well. 
In summary, we make the following contributions: 
\begin{itemize}
    \item  A global and focal architecture that effectively expands NeRF capacity while maintaining global consistency.
    \item  A global-guided training strategy that fully utilizes the rich priors from the global stage.
    \item  Experiments on large street and aerial-view scenes show GF-NeRF can achieve high-quality renderings.
\end{itemize}



\section{Related Works}
\subsection{Neural Radiance Field}
The neural radiance field is proposed by Mildenhall et al. \cite{mildenhall2021nerf}  as an implicit neural representation for novel view synthesis. With the positions and directions of 3D points as input, NeRF models the colors and volume densities corresponding to the points through a multi-layer perceptron (MLP), forming a hidden representation of 3D scene. Various types of NeRFs have been proposed to improve original NeRF \cite{barron2021mip,martin2021nerf,gao2022nerf,deng2022depth}. Researchers have shown that the computationally expensive MLPs in original NeRF are not necessary and proposed grid-based methods to store scenes' features and accelerate training and rendering \cite{yu2021plenoctrees,muller2022instant,fridovich2022plenoxels,chan2022efficient}. Plenoxels proposes to employ a voxel grid to learn 3D scenes' color and density \cite{fridovich2022plenoxels}. Instant-NGP adopts a hash table to encode scenes' features \cite{muller2022instant}, its efficient CUDA implementation greatly accelerates NeRF training and rendering. F2-NeRF \cite{wang2023f2} takes a further step to improve performance with an anchored hash table. Therefore, in this work, we use the anchored hash table in F2-NeRF as the base encoder of the scene for fast rendering and better performance.
\subsection{Large-Scale NeRF}
Since the emergence of NeRF, numerous works have sought to extend its application to large-scale scenes. For instance, Block-NeRF \cite{tancik2022block} divides street scenes into multiple blocks based on geographical location and exploits reciprocal distances as the weights to blend blocks' outputs. Block-NeRF enables fly-throughs in expansive city scenes, however, its division method relies heavily on strong priors, and the blending introduces blurred results in blocks' neighbored areas. Another line of research focuses on reconstructing and rendering large-scale scenes from aerial or satellite images\cite{turki2022mega,zhenxing2022switch,xu2023grid,xiangli2022bungeenerf}. Most of them introduce diverse designs based on the characteristics of the target datasets, such as ellipsoidal sampling \cite{turki2022mega}, mixture of experts \cite{zhenxing2022switch}, residual structures\cite{xiangli2022bungeenerf}, addressing issues of training and capacity expansion in such scenes. While these works have achieved impressive results in aerial/satellite images, their highly customized designs are difficult to extend to other types of scenes, such as street-view scenes. While GF-NeRF still follow the divide-and-conquer paradigm similar to \cite{tancik2022block,turki2022mega} to expand model capacity, it avoids the defects of requiring partitioning priors and inconsistencies between blocks. Furthermore, in comparison to methods targeting aerial/satellite datasets, our approach can handle various types of large-scale scenes, such as street-view and aerial-view scenes.  Recent advances in 3D gaussian splattin(3DGS) \cite{kerbl20233d} that model scenes from point clouds have demonstrated superior performance over NeRF in object-level rendering tasks due to their exceptional rendering speed and realistic results. However, the significantly increasing number of point clouds in large-scale scenes renders 3DGS impractical in terms of memory consumption. In contrast, GF-NeRF decouples scene size from memory usage, enabling the training of large-scale scenes on a single consumer-level GPU.

\section{Method}

Recall that the divide-and-conquer paradigm divides scenes into multiple smaller, mutually independent blocks, allowing NeRF to effectively reconstruct large-scale scenes. However, due to the independent training of each block, there are notable inconsistencies in geometry and appearance among them. On the other hand, the coarse-to-fine paradigm progressively refines scene details, inherently ensuring global consistency, but does not directly increase the model capacity and the prior information of the coarse stage has not been fully exploited.
\begin{figure*}[t]
\centering 
\includegraphics[width=0.8\linewidth]{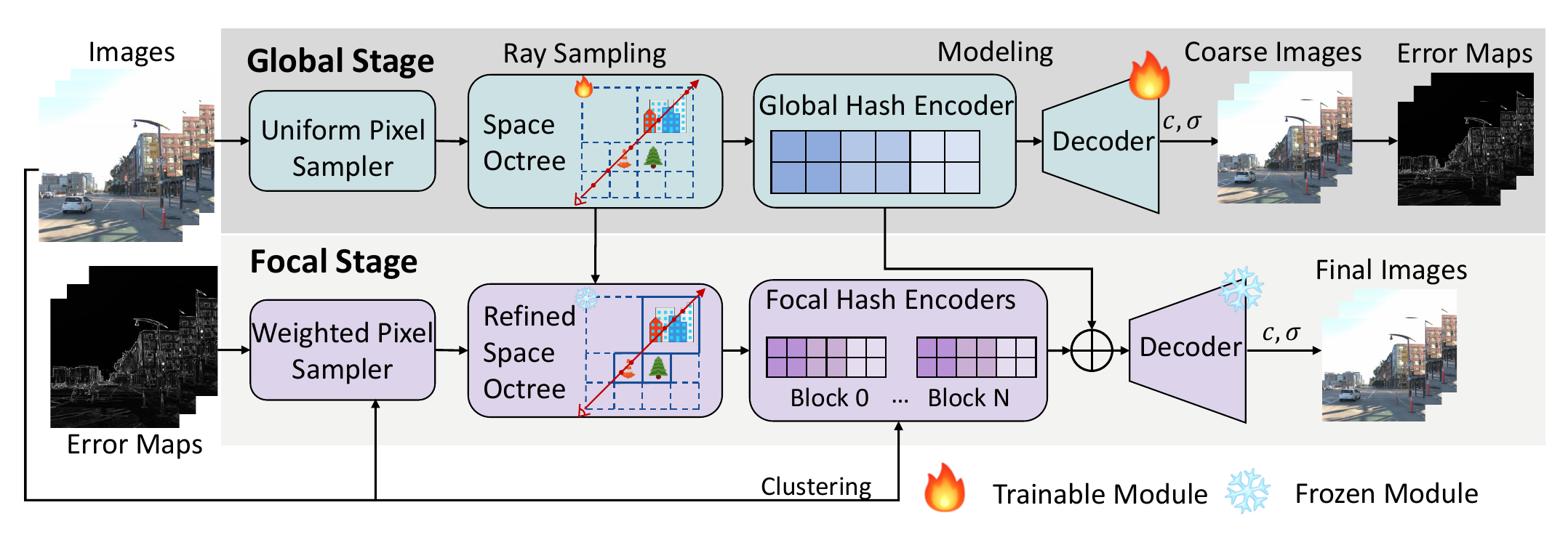} 
\caption{Overview of our framework. A two-stage architecture is adopted to model large-scale scenes. Global stage aims to learn target scene's coarse representation and focal stage refines high-frequency details with the guidance from global stage.} 
\label{fig:overview} 
\end{figure*}

To better represent large-scale scenes, we propose the global-guided focal neural radiance field, which integrates the strengths of the above two paradigms. As illustrated in Fig.~\ref{fig:overview}, our proposed framework employs a two-stage strategy consisting of a Global stage and a focal stage for efficient NeRF training on large-scale scales. In the global stage, the model is trained on the entire dataset to capture a holistic, continuous, and consistent coarse representation of the target scene. Subsequently, the scene is divided into blocks where the focal stage focuses on capturing the high-frequency details. In addition, the global prior information about the scene in the global stage can be further used to guide the learning of focal stage.

Within our framework, we utilize F2-NeRF~\cite{wang2023f2} as the base NeRF model, which optimally exploits a hash table encoder to retain an extensive amount of features. Comprehensive details regarding NeRF and F2-NeRF will be presented in Section~\ref{sec:preliminaries}. Each module within the global and focal stages will be introduced in Section~\ref{sec:model}. Finally, we will provide the loss function and block splitting details in Section~\ref{sec:training}.

\subsection{Preliminaries on NeRF and F2-NeRF}\label{sec:preliminaries}
\textbf{NeRF.} Neural radiance field \cite{mildenhall2021nerf} parameterizes scenes into volumetric density and color fields from posed images. The color of each pixel in an image is attributed to the accumulation of corresponding ray within the 3D fields.  Assuming a ray in 3D space is $\mathbf{r}(t) = \mathbf{o} + t\mathbf{d}$, where $\mathbf{o}$ is the origin and $\mathbf{d}$ is the direction, the sampled points along this ray are $\mathbf{r}(t_i)$, where $\{t_i\}_{i=0}^{N}$ is the set of sampled distances. NeRF first inputs the sampled points and directions into the volumetric fields to obtain $\{c_i\}_{i=0}^{N}$ and $\{\sigma_i\}_{i=0}^{N}$, and then accumulates them using volume rendering:

\begin{equation}
    C_{out} = \sum_{i=1}^{N}w_ic_i, \quad where \ w_i = T_i(1-e^{-\Delta_i\sigma_i})
\end{equation}
\begin{equation}
    T_i = \exp(-\sum_{j<i}\Delta_j\sigma_j), \quad\Delta_i = t_i - t_{i-1}
\end{equation}
$C_{out}$ is the final output color. Finally, NeRF optimizes its 3D fields by minimizing the loss between $C_{out}$ and the ground truth color in the training images.

\noindent\textbf{F2-NeRF.} 
The original NeRF utilizes MLPs as the representation of the density and color fields, resulting in a significant computation cost. To address this issue, hash table encoder \cite{muller2022instant} is proposed to store scene features, and only tiny MLPs are required as decoder to parse features into density and color, therefore reducing query time cost. F2-NeRF \cite{wang2023f2} takes a further step by introducing the anchored hash table, which employs different hash functions to perform table look-ups. F2-NeRF first subdivides the original Euclidean space and computes their perspective transformation matrices $\{M_i\}_{i=0}^N$. Then each region in the original space is projected into the same local space:
\begin{equation}
    \mathbf{z}_j = M_i(\mathbf{x}_j) \quad  \mathbf{x}_j\in \mathbb{R}^3, \ \mathbf{z}_j \in \mathbb{R}^3
\end{equation}
where $\mathbf{x}_j$ is a 3D point in the original Euclidean space and $\mathbf{z}_j$ is the warped point in local space. Finally, points in local space are looked up in the anchored hash table through different hash functions:
\begin{equation}
    Hash_i(\mathbf{z}) = (\bigoplus_{k=1}^3 z_k\pi_{i,k}+b_{i,k}) \mod L
\end{equation}
where $z_k$ is coordinate of the point in local space and $k=1,2,3$ corresponds to $x,y,z$. $\bigoplus$ denotes the bit-wise XOR operation, and $\{\pi_{i,k}\}$ and $\{b_{i,k}\}$ are large prime numbers, which are unique for each region in original space. $L$ is the length of the anchored hash table. Each region in the original space is mapped to the local space, meaning that every point in the original space has the opportunity to be mapped to any location in the hash table. Therefore, the capacity of the anchored hash table can be fully utilized.

\subsection{Global-guided Focal Model}\label{sec:model}
As shown in Fig.\ref{fig:overview}, we divide the training of GF-NeRF into {global} and {focal} stages. The former captures the global information of the target scene, and the latter further reconstructs the details of the sub-scene. Both stages employ an analogous pipeline consisting of three modules: pixel sampling, ray sampling, and modeling with the encoder and decoder. The focal stage capitalizes on the global priors established in the global stage, facilitating a more efficient training process. To elucidate the interplay between global and focal stages, we provide an in-depth analysis and comparative study of the three modules within each stage.

\begin{figure}[t]
\centering 
\includegraphics[width=1.0\linewidth]{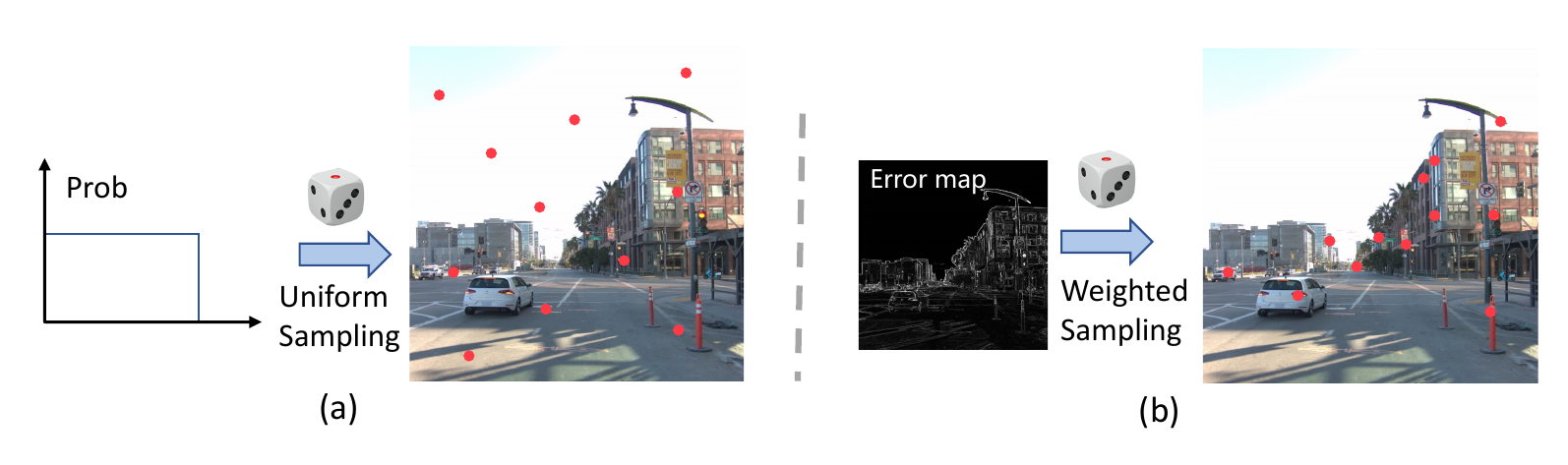} 
\caption{Different pixel sampling strategy. (a) Uniform pixel sampling that used in global stage. (b) Weighted pixel sampling that used in focal stage, enabling the sampling points to focus on the areas that did not perform well in global stage.} 
\label{fig:pixel-sampler} 
\end{figure}

\noindent \textbf{Pixel Sampling.} The role of pixel sampling is to sample pixels on the image plane for training. In the original F2-NeRF, uniform pixel sampling across all training images is employed, which is instrumental in capturing the scene's global structure and steering toward optimal reconstruction. In line with this, our global stage also utilizes uniform pixel sampling as depicted in Fig. \ref{fig:pixel-sampler} (a). Conversely, the focal stage is designed to refine the details overlooked by the global stage, necessitating a more targeted sampling strategy. Hence, we introduce weighted pixel sampling, as illustrated in Fig. \ref{fig:pixel-sampler} (b). This advanced sampling method leverages the global stage's output to dynamically guide the pixel sampling process, thereby enhancing the focal stage's training efficiency. Specifically, we generate coarse re-renderings of all training images via the global stage, namely coarse images $\{\hat{I}_i\}_{i=1}^{m}$. Subsequently, we compute the Mean Absolute Error~(MAE) between these coarse images and their corresponding ground-truth images $\{I_i\}_{i=1}^{m}$, yielding a set of error maps 
$\{Err_i\}_{i=1}^{m}$, one for each training image: 
 \begin{equation}
 Err_i = \Vert \hat{I}_i - I_i \Vert_1
 \end{equation}

 Considering the impracticality of rendering high-resolution images for large-scale datasets, we opt for low-resolution rendering followed by upscaling to the original size. Since the error map serves exclusively as a weighting mechanism for pixel sampling, its upscaling error has no significant impact on the final reconstruction. The weighted pixel sampling calculates the sampling probability for each pixel during each epoch, based on the MAE values across all pixels within the sampled images:
 \begin{equation}
 p_j = \frac{e_j}{\sum_{k=1}^n e_k}
 \end{equation}
where $e_j$ is the MAE value of the pixel with index of $j$, and $n$ is the total number of pixels in a batch. 

As training advances, the focal stage dynamically identifies areas requiring further refinement. However, relying solely on error map based sampling risks neglecting well-rendered regions in the global stage, such as the sky or ground. To counteract this, we complement weighted pixel sampling with uniform pixel sampling for a subset of the pixels. This hybrid approach, guided by the global stage, enables GF-NeRF to efficiently learn scene details and mitigates training inefficiencies associated with the large-scale scene.

\noindent \textbf{Ray Sampling.} Ray sampling, distinct from pixel sampling, involves the selection of points along a ray at various intervals. Unlike the original NeRF's hierarchical sampling strategy, which concentrates more sampling points in areas of the scene with higher density~(such as object surfaces), our framework employs the ray sampling approach similar to F2-NeRF that leverages a spatial octree to circumvent the redundancy of hierarchical sampling by directly querying the octree. As F2-NeRF points out, this sampling method makes the distribution of sampling points in the scene more reasonable, helping to capture the details of the scene.

During the global stage, the spatial octree undergoes dynamic updates based on the cumulative density of each region. Regions falling below a predefined threshold are pruned, while denser areas are subdivided, allowing the ray sampling to efficiently navigate through the scene's surface while bypassing empty spaces. The resulting refined octree from the global stage captures a comprehensive and contiguous representation of the scene's occupancy.

For the focal stage, we maintain the refined octree, utilizing it directly for ray sampling and no longer updating it. The reason for this is that the focal stage's training operates on sub-datasets, altering the octree in one block will disrupt the ray sampling process for other blocks. By freezing the octree, we ensure a consistent and coherent sampling strategy across the entire scene.


\subsection{Comparison with Baselines}

\noindent \textbf{Modeling.} 
The primary obstacle in deploying NeRF for large-scale scenes lies in the model's capacity constraints. To overcome this, we split the entire scene into discrete blocks, each equipped with its own hash encoder to store localized features. Although Block-NeRF has employed a similar strategy to enhance NeRF's capacity, it typically trains each sub-NeRF independently, starting from scratch. This approach not only complicates the training process but also introduces discrepancies in geometry and appearance across the sub-NeRFs. To address these issues, our method utilizes a global encoder to extract coarse features, which then guides the focal encoder to refine and learn the residual features. This collaborative encoding strategy ensures a cohesive and consistent representation across the entire scene.


The global hash encoder, as detailed in Sec. \ref{sec:preliminaries}, is a fixed-sized anchored hash table. This encoder takes 3D points as input, accompanied by the index of the octree node that these points are associated with. Subsequently, it employs a specific hash function $Hash_i(\cdot)$ corresponding to node index $i$ to look up point $\mathbf{z}$ in the table and output the feature $F_g$:
\begin{equation}
    F_g = E_g(Hash_i(\mathbf{z}))
\end{equation}
However, for large-scale scenes, a fixed-size hash table is usually insufficient and may lead to hash collisions. These conflicts arise when distinct 3D points are inadvertently assigned to the same hash table entry, which can significantly affect the accuracy of the stored feature. Consequently, the global hash encoder can only capture a coarse-grained representation of the scene. To complement the fine-grained representation of the scene, we introduce the focal hash encoder for each local block at the focal stage. Specifically, each local block is assigned a dedicated focal hash encoder, which is responsible for retrieving features that were not captured during the global stage. We treat the focal stage training as a meticulous fine-tuning process that builds upon the global stage. This approach is designed to maintain global consistency while incorporating the unique details contributed by each block. To this end, we immobilize the global hash encoder to direct the focal encoder to concentrate on refining and learning the residual features. The hash features produced by the focal stage, denoted as $F_o$, are articulated as follows:


\begin{equation}
    F_o= E_g(Hash_i(\mathbf{z})) + E_f(Hash_i(\mathbf{z}))
\end{equation}
where $E_f$ is the focal hash encoder. To avoid significant bias during the initial training phase in focal stage, we initialize focal hash encoder to zero, ensuring that the focal stage fine-tunes based on the global output. We perform fusion operation at the hash-feats level rather than density/color level like other methods\cite{xiangli2022bungeenerf,tancik2022block} since it is difficult to control the decoder to output zero density/color.

Another mechanism employed to maintain consistency lies within the decoder. The decoder is responsible for parsing the corresponding density $\sigma$ and color $c$ from the hash features $F$, i.e.,
\begin{equation}
    \sigma,c= D(F_g,\mathbf{d})
\end{equation}
where $\mathbf{d}$ is direction of the ray. The decoder consists of two tiny MLPs and their weights are updated only during the global stage. Employing a shared decoder helps alleviate the problem of inconsistent appearances across different blocks.

Above design guarantees that the training for each block starts from the result of the {global} stage, thereby significantly alleviating the training complexity in {focal} stage. Moreover, fine-tuning each block based on the global results contributes to the geometry and appearance consistency across blocks in GF-NeRF.

\subsection{Training}\label{sec:training}

The loss function for training GF-NeRF is defined as
\begin{equation}
	\mathcal{L} = \sqrt{(C_{out} - C_{gt})^2 + \epsilon}
 \end{equation}
 where $\mathcal{L}$ is a Charbonnier Loss \cite{barron2022mip} used for color reconstruction and the default value of $\epsilon$ is $10^{-6}$. We use the same loss function in both global and focal stages.
 
 {Global} stage is trained on all images of the training dataset while {focal} stage exploits sub-datasets to train focal hash encoders individually to expand NeRF capacity and capture more details. We partition the training dataset into $k$ sub-datasets based on the positions of the cameras and utilize separate hash table encoders to store detailed features within each block. 
 During evaluation, we calculate the distance from the camera position to the center of each block and activate the nearest block for subsequent rendering. Since the blocks in focal stage are independent of each other, they can be trained serially on a single GPU or simultaneously trained on multiple GPUs.
 \begin{table*}[t]
  \centering
  \caption{Quantitative comparison results of aerial-view baselines and our method. The best and second-best results in the table are marked as (\textbf{best}/\underline{second best}).}
  \resizebox{\linewidth}{!}{
    \begin{tabular}{c|ccc|ccc|ccc|ccc|ccc}
    \toprule
    Dataset & \multicolumn{3}{c|}{Campus} &       & Residence &       & \multicolumn{3}{c|}{Sci-Art} & \multicolumn{3}{c|}{Building} & \multicolumn{3}{c}{Rubble} \\
    \midrule
    Metric & PSNR$\uparrow$ & SSIM$\uparrow$ & LPIPS$\downarrow$ & PSNR$\uparrow$ & SSIM$\uparrow$ & LPIPS$\downarrow$ & PSNR$\uparrow$ & SSIM$\uparrow$ & LPIPS$\downarrow$ & PSNR$\uparrow$ & SSIM$\uparrow$ & LPIPS$\downarrow$ & PSNR$\uparrow$ & SSIM$\uparrow$ & LPIPS$\downarrow$ \\
    \midrule
    Mega-NeRF & 23.42 & 0.537 & 0.618 & 22.57 & 0.628 & 0.489 & 25.60 & 0.770 & 0.390 & 20.93 & 0.547 & 0.504 & 24.06 & 0.553 & 0.516 \\
    Switch-NeRF & \underline{23.62} & \underline{0.541} & \underline{0.609} & \underline{22.57} & \underline{0.654} & \underline{0.457} & \underline{26.52} & \underline{0.795} & \underline{0.360} & \underline{21.54} & \underline{0.579} & \underline{0.474} & \underline{24.31} & 0.562 & 0.496 \\
    3DGS  & failed & fail  & fail  & 21.7  & 0.814 & 0.221 & fail  & fail  & fail  & 20.81 & 0.565 & 0.483 & 24.08 & \underline{0.563} & \underline{0.497} \\
    Ours  & \textbf{24.38} & \textbf{0.640} & \textbf{0.375} & \textbf{24.02} & \textbf{0.760} & \textbf{0.203} & \textbf{28.37} & \textbf{0.855} & \textbf{0.158} & \textbf{22.61} & \textbf{0.635} & \textbf{0.349} & \textbf{26.37} & \textbf{0.684} & \textbf{0.284} \\
    \bottomrule
    \end{tabular}%
    }
  \label{tab:baseline-aerial}%
\end{table*}%
\begin{figure*}[t]
\centering 
\includegraphics[width=\linewidth]{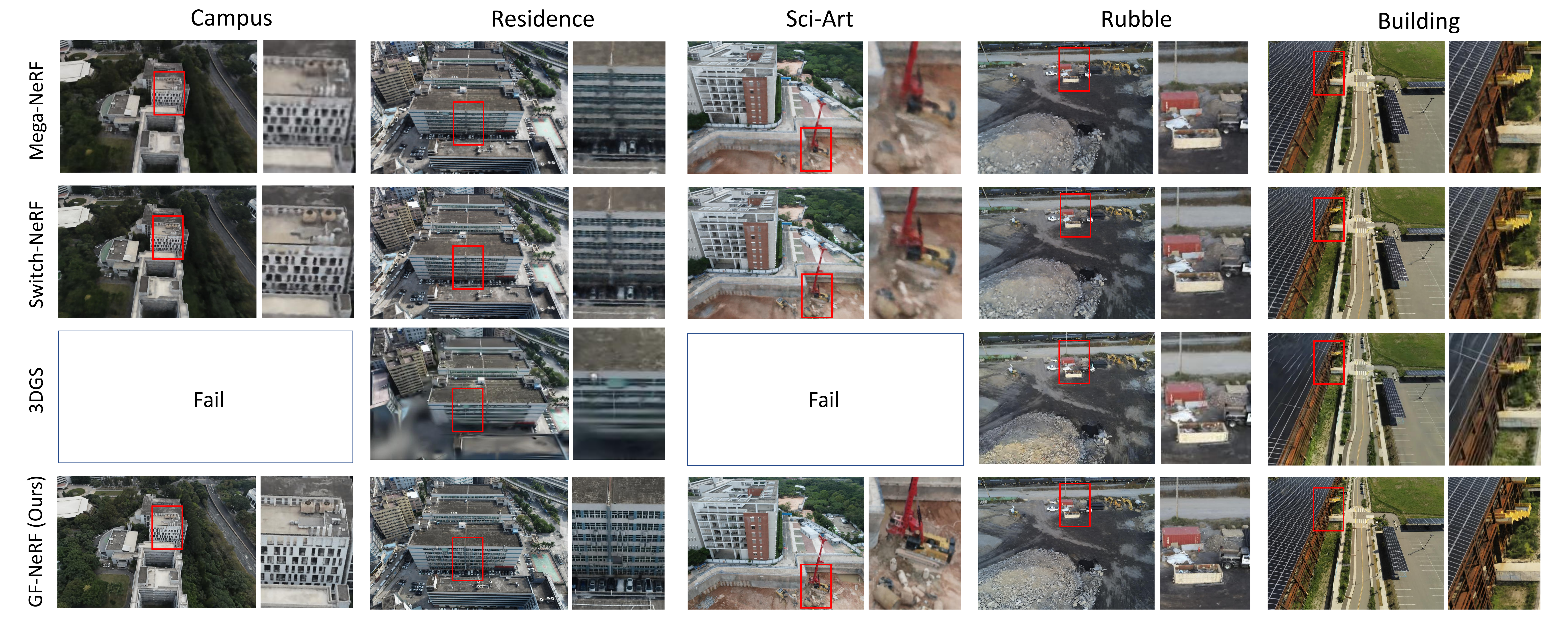} 
\caption{Qualitative comparison results on aerial scenes. Our method demonstrates superior detail rendering compared to Mega-NeRF and Switch-NeRF. We encourage readers to zoom in for a detailed visual comparison. } 
\label{fig:baseline-aerial} 
\end{figure*}

\section{Experiments}

The training of GF-NeRF is not contingent upon any prior assumptions regarding the scene, thus enabling its applicability across datasets collected through various acquisition approaches. Therefore, we conduct experiments on both large-scale aerial-view and street-view scenes. We compare our method to Mega-NeRF \cite{turki2022mega}, Switch-NeRF \cite{zhenxing2022switch} on aerials scenes, Block-NeRF \cite{tancik2022block}, F2-NeRF \cite{wang2023f2} on street scenes and 3DGS \cite{kerbl20233d} on both scenes. Additionally, we delve into a thorough analysis of the effects of our global-guided focal framework through ablation studies.

\noindent\textbf{Datasets.}
We evaluate our GF-NeRF on two types of datasets: aerial-view and street-view. For aerial-view, we use Campus, Sci-Art, and Residence datasets from UrbanScene3D \cite{lin2022capturing} and Rubble, Building datasets from Mill19 \cite{turki2022mega}. These datasets have the same settings with Mega-NeRF \cite{turki2022mega}, including camera poses and train-test splitting. For street view evaluation, we opt for a real-world dataset San Francisco Misson Bay dataset \cite{tancik2022block} and two synthetic datasets Block\_A (5m interval), Block\_Small (0.5 m interval) from MatrixCity \cite{li2023matrixcity}. Additional details regarding these street datasets are provided in the supplementary material.

\noindent \textbf{Metrics.} We use the metrics PSNR, SSIM \cite{wang2004image} for low-level quantitative comparison and the VGG implementation of LPIPS \cite{zhang2018unreasonable} (lower is better) for perceptual similarity measurement. 

\noindent \textbf{Implementation.} GF-NeRF is implemented on Nerfstudio \cite{tancik2023nerfstudio} and trained on a single NVIDIA A100-80G GPU. Each batch comprises $8192$ sampled rays with a maximum of $1024$ points per ray. We employ the Adam optimizer \cite{kingma2014adam} with an initial learning rate of $1e^{-2}$ decaying exponentially to $1e^{-4}$ in global stage, while the learning rate in focal stage is decaying from $5e^{-3}$ to $5e^{-5}$. The optimizer is reset at the commencement of each new stage or new block training. To reduce memory consumption, we downsample all training images by a factor of $2$. 

\begin{table*}[htbp]
  \centering
  \caption{Quantitative comparison results on large urban street scenes.}
  \resizebox{0.7\linewidth}{!}{
    \begin{tabular}{c|ccc|ccc|ccc}
    \toprule
    Dataset & \multicolumn{3}{c|}{Misson Bay} & \multicolumn{3}{c|}{Block\_Small} & \multicolumn{3}{c}{Block\_A} \\
    \midrule
    Metric & PSNR$\uparrow$ & SSIM$\uparrow$ & LPIPS$\downarrow$ & PSNR$\uparrow$ & SSIM$\uparrow$ & LPIPS$\downarrow$ & PSNR$\uparrow$ & SSIM$\uparrow$ & LPIPS$\downarrow$ \\
    \midrule
    F2-NeRF & \underline{23.94} & 0.785 & 0.264 & 23.58 & 0.770 & 0.263 & 20.23 & 0.601 & 0.413 \\
    Block-NeRF & 23.56 & \underline{0.812} & \underline{0.221} & \underline{23.85} & \underline{0.786} & \underline{0.221} & \underline{21.24} & \underline{0.695} & \underline{0.358} \\
    3DGS & Fail & Fail & Fail & {21.42} & {0.642} & {0.308} & {Fail} & {Fail} & {Fail} \\

    Ours  & \textbf{25.08} & \textbf{0.825} & \textbf{0.186} & \textbf{24.85} & \textbf{0.841} & \textbf{0.157} & \textbf{23.51} & \textbf{0.730} & \textbf{0.287} \\
    \bottomrule
    \end{tabular}%
    }
  \label{tab:baseline-street}%
\end{table*}%
\begin{figure*}[htbp]
\centering 
\includegraphics[width=0.8\linewidth]{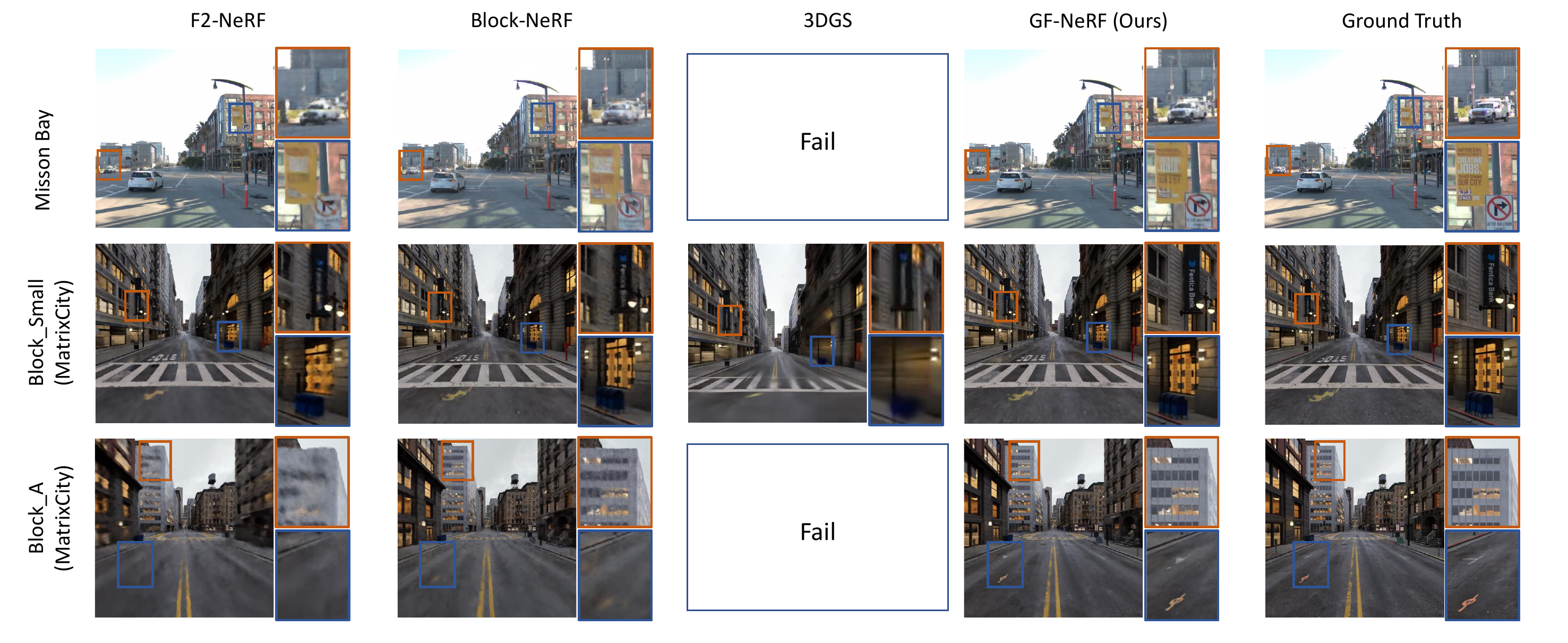} 
\caption{Qualitative comparison between baselines and our GF-NeRF on street scenes. F2-NeRF exhibits the most blurred renderings due to its limited model capacity. Block-NeRF manages to capture some details but still suffers from blurry results. In comparison, our method produces the highest rendering quality in street scenes.}
\label{fig:baseline-street} 
\end{figure*}

\noindent \textbf{Aerial Scenes.} 
The visualization of the comparison is depicted in Fig. \ref{fig:baseline-aerial}. While Mega-NeRF has made specific improvements for aerial scenes, its rendering results appear to be over-smoothed. Switch-NeRF incorporation a learnable decomposition technique to improve upon Mega-NeRF, but it still struggles to reveal the fine structures in high-frequency areas, such as windows and building trusses. The recently popular 3DGS method often fail to optimize in scenarios with sparse views, such as Block\_A and Waymo, due to the low-quality initialization of point clouds. Additionally, in dense scenes, the rendering details remain suboptimal because of the limitations in the number of point clouds. With our global-guided training strategy, GF-NeRF can notice and capture these fine structures even in such large scenes. As illustrated in Fig. \ref{fig:baseline-aerial}, our model renders more details in all scenes. The quantitative results in Table. \ref{tab:baseline-aerial} further substantiate the superiority of our method over baselines on aerial scenes.

\begin{figure}[htbp]
\centering 
\includegraphics[width=1.1\linewidth]{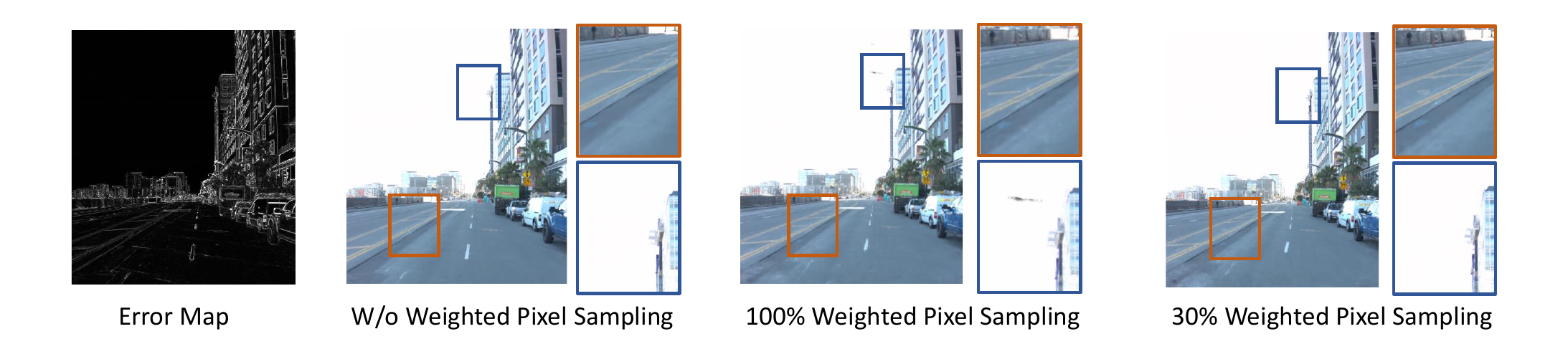} 
\caption{Qualitative ablation results of weighted pixel sampling. $100\%$ represents all pixels are sampled from weighted pixel sampling, while $30\%$ means only $30\%$ of pixels come from weighted pixel sampling and the other pixels are sampled uniformly. }  
\label{fig:importance-pixel} 
\end{figure}
\noindent \textbf{Street Scenes.} We conduct comparisons between our method and Block-NeRF, and F2-NeRF on street scenes. In Block-NeRF, we adopt the same number of blocks as GF-NeRF. For F2-NeRF, we employ a hash table with a size of $2^{21}$, while keeping other parameters consistent with GF-NeRF. The qualitative and quantitative results are reported in Fig. \ref{fig:baseline-street} and Table. \ref{tab:baseline-street}, respectively. Street scenes inherently contain more details than aerial scenes, which raises challenges for model capacity, thereby F2-NeRF exhibits the most blurred results with a limited model capacity. Similar to its performance in aerial scenes, 3DGS exhibits a high failure rate in street scenes. Even in successfully reconstructed scene, the rendering results tend to be blurry due to the limitations in the number of point clouds.


Compared to F2NeRF, Block-NeRF produces sharper results, indicative of its enhanced model capacity that retains certain high-frequency details. Nonetheless, disparities between renderings and ground-truths persist, particularly evident in the Misson Bay dataset characterized by significant exposure variations. Block-NeRF composites colors from different blocks, resulting in variation of appearance and consequent blurring results. In contrast, our GF-NeRF trains focal blocks based on the coarse representation from global stage, expanding capacity while avoiding explicit compositing operations. Hence, GF-NeRF exhibits a superior ability to preserve details in street scenes compared to the baselines.

\begin{figure}[htbp]
\centering 
\includegraphics[width=0.7\linewidth]{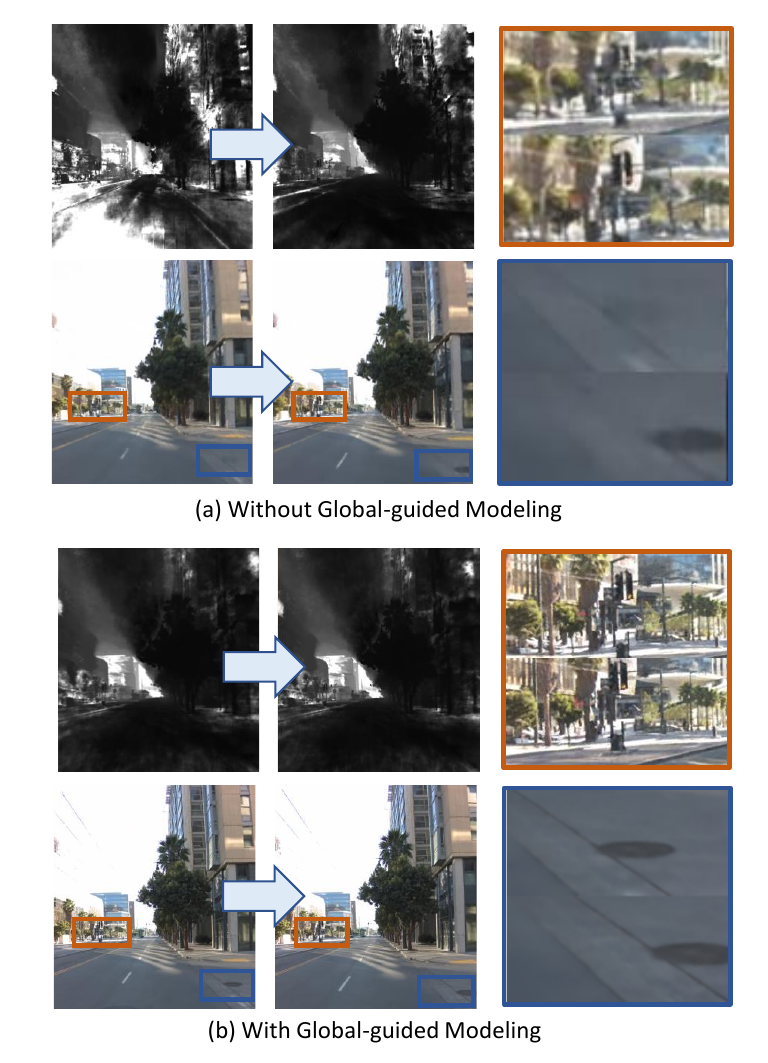} 
\caption{Visualization results of the ablation study on global-guided modeling. The two images on the left in the first row are depth images.}  
\label{fig:importance-modeling} 
\end{figure}

\subsection{Ablation Studies}
In addition to comparing with baselines, we conducted ablation experiments on key modules in GF-NeRF, including global-guided modeling and weighted pixel sampling. To clearly illustrate the effectiveness of these modules, we perform experiments on the Mission Bay dataset, which features rich details and challenging appearance variations.


\noindent \textbf{Effectiveness of Global-guided Modeling.}
As mentioned in Sec. \ref{sec:model}, we regard the training process of focal stage as a fine-tuning procedure built upon global stage. To demonstrate the effectiveness of this global-guided modeling, we remove the guidance from the global encoder. Specifically, we retain the global stage to ensure fair pixel sampling and ray sampling while the hash features required for predicting density and color in the focal stage are solely provided by the focal hash encoders. This means that each block's encoder in focal stage is trained from scratch independently. The quantitative results of the comparison are presented in Table .\ref{tab:ablation}, indicating that removing global-guided modeling results in a decrease across all metrics. 

In Fig.\ref{fig:importance-modeling} (a), we visualize the consistency of the geometry and appearance between two images from adjacent blocks after removing the global guidance. It is evident that although the contents of these two images are very similar, there are significant disparities in the geometry reflected in their depth maps, and the geometric inconsistency further exacerbates the differences in appearance. In contrast, Fig. \ref{fig:importance-modeling} (b) presents the results with global-guided modeling, where the geometry and appearance of images from different blocks exhibit high consistency. Furthermore, global-guided modeling significantly improves rendering quality, indicating that training from the global facilitates convergence compared to training from scratch. Above ablation results demonstrate the effectiveness of global-guided modeling in mitigating training complexity, expanding model capacity, while simultaneously preserving consistency in geometry and appearance.

\begin{table}[htbp]
  \centering
  \caption{Quantitative ablation study results on global-guided modeling and weighted pixel sampling.}
    \resizebox{\linewidth}{!}{
    \begin{tabular}{c|c|c|c|c}
    \toprule
    Metric & \makecell{W/o Global-guided \\ Modeling} & \makecell{W/o Weighted \\  Pixel Sampling} & \makecell{$100\%$ Weighted \\ Pixel Sampling} & \makecell{$30\%$ Weighted \\ Pixel Sampling (Ours)} \\
    \midrule
    PSNR$\uparrow$ & 24.45 & 24.58 & \underline{24.66} & \textbf{25.08} \\
    SSIM$\uparrow$ & 0.803 & 0.805 & \underline{0.810} & \textbf{0.825} \\
    LPIPS$\downarrow$ & 0.231 & \underline{0.190} & 0.191 & \textbf{0.186} \\
    \bottomrule
    \end{tabular}%
    }
  \label{tab:ablation}%
\end{table}%

\noindent \textbf{Effectiveness of Weighted Pixel Sampling.} To illustrate the effectiveness of the weighted pixel sampling, we conduct comparison experiments on uniform pixel sampling (w/o weighted pixel sampling) and weighted pixel sampling with different proportions. The qualitative and quantitative results are presented in Fig. \ref{fig:importance-pixel} and Table. \ref{tab:ablation}. The results reveal that the results with weighted pixel sampling are more sharper than pure uniform sampling, especially in areas where values in error maps are high. This suggests that the error map generated from global stage effectively guides the focal stage to focus on areas with suboptimal performance and capture finer details.

However, excessive reliance on weighted pixel sampling will result in additional issue. As illustrated in Fig. \ref{fig:importance-pixel}, when all pixels within a batch are sampled from weighted pixel sampling, although more details are revealed on the road surface (orange boxes), the original clean sky exhibits noisy artifacts (blue boxes). This phenomenon arises since the global stage performs well in low-frequency areas such as the sky, the focal stage will overlook these areas, resulting in tiny or zero gradient accumulation during training. As a consequence, these areas without gradient retention are at a disadvantage in hash collision competition, leading to the emergence of noisy artifacts. To alleviate this issue, we adopt a mixed sampling strategy. For instance, $30\%$ of pixels are sampled with error guidance, while the remaining pixels are uniformly sampled, hence achieving a balance between detail preservation and noise reduction.

\section{Discussion and Conclusion}
In this work, we introduce GF-NeRF, a global-guided focal neural radiance field tailored for rendering large-scale scenes. We divide the training of large-scale NeRF into two stages and GF-NeRF utilizes rich priors obtained from the global stage about the entire scene to guide the training process of each block in the focal stage. The integration of global and focal stages enables GF-NeRF to maintain geometric and appearance consistency while expanding model capacity. Additionally, our method can focus on important regions to capture more intricate details. 

Despite achieving high-fidelity rendering results on various types of large-scale datasets, there are still some challenges that we aim to address in the future: (1) GF-NeRF's training and rendering speeds are still relatively slower compared to current fastest rendering methods, such as 3D gaussian splatting \cite{kerbl20233d}. (2) While we decouple the memory consumption with the number of hash encoders, in extremely large scenes, the memory usage for the space octree can not be ignored.

{\small
\bibliographystyle{ieee_fullname}
\bibliography{egbib}
}
\clearpage
\appendix
\section{Dataset Deails}
In our main manuscript, we employ three three street-view datasets: San Francisco Misson Bay, Block\_small (MatrixCity), and Block\_A (MatrixCity) to evaluate our method performance in street scenes. Here we introduce the details of these three street datasets.
\subsection{San Francisco Misson Bay}
The San Francisco Misson Bay dataset is a street view dataset proposed by Block-NeRF~\cite{tancik2022block}. This dataset is collected by an autonomous driving collection vehicle equipped with $12$ cameras and contains approximately $12k$ images from $1.2 km$ length street view. Different cameras in this public dataset are not accurately registered. In order to avoid the error from the unregistered cameras, we opt for the training images of the camera with camera ID of $\mathbf{69}$ among the $12$ cameras in the training set. Similarly, we filter out the images with camera ID of $\mathbf{69}$ from the test set for evaluation. It is worth noting that our setting in this dataset is only $1/12$ sparse compared to the original Block-NeRF paper, hence our metrics on the test set are lower than those reported in the original Block-NeRF.

\subsection{MatrixCity}
MatrixCity is a synthetic large-scale urban dataset built on Unreal Engine\cite{li2023matrixcity} where researchers of MatrixCity have constructed two cities in the virtual world: Big-City and Small-City, and have collected $67k$ aerial images and $452k$ street view images, covering a total urban area of $28km^2$. MatrixCity provided a street view benchmark for Block\_A and Block\_small in small cities, hence we also employ these two datasets for comparison. MatrixCity releases two versions of the two datasets, including the dense version ($0.5m$ sampling interval and the sparse version ($5m$ sampling interval). Since the Blcok\_A is a large street scene, a dense sampling with too many images ($20k$ images) would lead to an excessively long training time. Whereas, the Block\_small is relatively small, with only $1.5k$ images in the dense version. Thus, we opt for 0.5m interval version for Block\_small. All the metrics in our experiments are evaluated on the test views provided by MatrixCity.

\section{Detailed Configurations of Experiments}
We compare Mega-NeRF~\cite{turki2022mega} and Switch-NeRF~\cite{zhenxing2022switch} on aerial scenes. Since the code for both methods is publicly accessible, we directly utilize the checkpoints provided by these two methods to obtain the comparative results presented in our main document. As the scale of the aerial scenes in our experiment is roughly the same, we use identical configurations for our GF-NeRF across five aerial scenes, with specific settings detailed in Table \ref{tab:exp-configs} (GF-NeRF, Aerial-All).

On street-view scenes, we conduct comparisons between F2-NeRF \cite{wang2023f2}, Block-NeRF \cite{tancik2022block}, and GF-NeRF. We employ the official implementation of F2-NeRF, while for Block-NeRF since the code is not publicly available, we implement a version ourselves. Specifically, we implement Block-NeRF in nerfstudio using nerfacto as the base-NeRF~\cite{tancik2023nerfstudio}. Unlike Mip-NeRF~\cite{barron2021mip}, nerfacto employs a hash table to store scene features instead of MLP. As MatrixCity lacks explicit geographical data for block partitioning, in our self-implemented Block-NeRF, we adopt the same block partitioning method as our GF-NeRF, i.e., balanced clustering algorithm. To ensure consistency among blocks during rendering, we calculate the nearest three blocks based on the rendering camera position and render them separately. Finally, we perform weighted compositing based on the reciprocal of the distances to the centers of these three blocks to get the final color.

The geographical extents covered by the three street-view scenes selected in our paper vary significantly. Hence, we use different configurations for each scene, with specific parameters outlined in Table. \ref{tab:exp-configs}. As the scene Block-A covers the largest area, we utilize a large number of blocks, hash table size, and batch size to ensure that NeRF's capacity is sufficient to store the entire scene and can be adequately trained. Ablation experiments are conducted on the Misson Bay Dataset, with all parameters except the ablation module being consistent with those listed in Table. \ref{tab:exp-configs}.

\begin{table}
  \centering
  \caption{Detailed configurations of compared methods on different scenes.}
  \resizebox{\linewidth}{!}{
    \begin{tabular}{c|c|c|c|c|c|c}
    \toprule
    Method & Dataset & \makecell{Number of \\ Blocks}& Global Steps & \makecell{Focal (Block) \\ Steps} & Batch Size & \multicolumn{1}{c|}{\makecell{Hash Table Size \\(log2)}} \\
    \midrule
    \multirow{3}[2]{*}{F2-NeRF} & Street-Misson & /     & 400k  & /     & 8192  & 21 \\
          & Street-Block\_A & /     & 500k  & /     & 16384 & 23 \\
          & Street-Block\_small & /     & 400k  & /     & 8192  & 21 \\
    \midrule
    \multirow{3}[2]{*}{Block-NeRF} & Street-Misson & 10    & /     & 30k   & 8192  & 21 \\
          & Street-Block\_A & 15    & /     & 30k   & 16384 & 23 \\
          & Street-Block\_small & 10    & /     & 30k   & 8192  & 21 \\
    \midrule
    \multirow{4}[2]{*}{\makecell{GF-NeRF}} & Aerials-All & 15    & 100k  & 30k   & 8192  & 21 \\
          & Street-Misson & 10    & 100k  & 30k   & 8192  & 21 \\
          & Street-Block\_A & 15    & 50k   & 30k   & 16384 & 23 \\
          & Street-Block\_small & 10    & 100k  & 30k   & 8192  & 21 \\
    \bottomrule
    \end{tabular}%
    }
  \label{tab:exp-configs}%
\end{table}%

\end{document}